\documentclass[11pt]{article}

\usepackage[margin=1in]{geometry}
\PassOptionsToPackage{numbers,sort&compress}{natbib}
\usepackage{natbib}
\usepackage[utf8]{inputenc}
\usepackage[T1]{fontenc}
\usepackage{amsmath,amssymb}
\usepackage{graphicx}
\usepackage{booktabs}
\usepackage{microtype}
\usepackage{float}
\usepackage{xcolor}
\usepackage[hidelinks]{hyperref}
\usepackage{url}

\title{Physics-Aware Auxiliary Losses Improve Out-of-Distribution\\
Generalization of a GNN Synthesizability Filter}

\author{%
  Riya Bisht\thanks{\texttt{manasi.riya2003@gmail.com}}
  \and
  Dhruv Agarwal\thanks{\texttt{dhruvagarwal5018@gmail.com}}
}

\date{May 28, 2026}

\begin{document}
\maketitle

\begin{abstract}
\noindent
Machine-learning drug-discovery pipelines increasingly rely on generative models
that propose molecules far from the data used to train downstream
\emph{synthesizability} filters. Existing filters (SAScore, SCScore, RAscore,
DeepSA) are purely statistical and degrade in exactly this out-of-distribution
(OOD) regime. We ask whether cheap, closed-form \emph{physical} priors, used as
auxiliary supervision on a graph neural network (GNN), improve OOD
generalization. We add two auxiliary losses to a GINE backbone: a topological
complexity regression supervised by the Bertz index, and a strain-energy soft
penalty supervised by MMFF94 force-field energy. On a 65{,}177-molecule corpus
(HIV, Tox21, COCONUT) labeled by SAScore thresholds we reproduce a strong
in-distribution baseline, then evaluate a 4-way ablation
(baseline / +complexity / +strain / +both) on a single-source OOD split
(train on drug-like HIV+Tox21, test on COCONUT natural products), repeated over
5 seeds with paired bootstrap confidence intervals. \textbf{All three
physics-aware variants give a small but statistically significant OOD
improvement} over the baseline (mean OOD AUC 0.9774): +complexity
$\Delta=+0.0060$ (95\% CI $[+0.0023,+0.0102]$), +strain $\Delta=+0.0032$
($[+0.0008,+0.0052]$), +both $\Delta=+0.0066$ ($[+0.0038,+0.0093]$); every
interval excludes zero, and the combination is best. The variants are
indistinguishable in-distribution, so the effect is visible only under OOD
evaluation. We are explicit that the effects are modest, and we report a
cautionary methodological finding: a single-seed version of this experiment
produced a qualitatively different (non-monotone) story that did not survive
multi-seed evaluation.
\end{abstract}

\section{Introduction}
Generative models for drug discovery---SMILES VAEs \citep{gomez2018}, graph
generative models, GFlowNet samplers \citep{bengio2021}, and graph diffusion
\citep{vignac2023}---produce candidate
molecules far faster than wet-lab validation can absorb. Every practical pipeline
therefore follows the generator with a cascade of cheap filters: predicted
binding affinity, ADMET property prediction, and---critically---predicted
\emph{synthesizability}. Without a synthesizability filter, generators routinely
propose strained, non-realizable substructures that no chemist can make in a lab,
no matter how strong their predicted binding.

The production-standard filters---SAScore \citep{ertl2009}, SCScore
\citep{coley2018}, RAscore \citep{thakkar2021}, and DeepSA \citep{deepsa2023}---are
statistical: each is fit to a labeled corpus of ``makeable'' versus ``hard''
molecules. They work near their training data and degrade outside it. This is the
single worst property a filter can have when it sits downstream of a model whose
\emph{job} is to produce novel structures: the filter is least reliable precisely
where the generator is most creative.

We take the following position:
\begin{quote}
\textit{Synthesizability prediction is a setting where physics-grounded inductive
bias belongs as a regularizer, not as an architectural backbone.} A GNN trained
jointly to predict synthesizability \emph{and} a small set of physically
meaningful quantities (strain energy, topological complexity) should learn
representations that transfer better to molecules outside its training
distribution than a purely statistical baseline.
\end{quote}
We commit in advance to the experimental protocol that tests this claim
(Sec.~\ref{sec:claims}) and report the result whether positive or null. Our
contributions are: (i) a clean formulation of two cheap, closed-form physics-aware
auxiliary losses on a standard GNN; (ii) a single-source OOD evaluation protocol
(a source-based covariate shift in the DrugOOD/GOOD sense) that isolates corpus
transfer for a synthesizability filter; (iii) a multi-seed, confidence-interval-based
result showing all three physics-aware variants give a small but significant OOD
improvement; and (iv) a documented, in-domain cautionary tale showing how a
single-seed run of an AI-assisted pipeline manufactured a false ``non-monotone''
narrative that multi-seed evaluation overturned.

\section{Why not a PINN}
Physics-informed neural networks (PINNs, \citealp{raissi2019}) assume (i) a known
governing PDE/ODE for the quantity of interest, (ii) a continuous domain over
which a neural field can be differentiated, and (iii) smooth dynamics so PDE
residuals are meaningful loss terms. Synthesizability satisfies none of these:
there is no governing equation (it depends on reagent availability, reaction
templates, kinetics, cost, and chemist judgment), the domain is a discrete
molecular graph rather than a continuous field, and bond formation is a discrete
graph rewrite rather than a smooth flow. The sound intuition---``use physics to
constrain learning''---therefore belongs as \emph{auxiliary losses on a graph
backbone}, not as a PDE-residual loss on a continuous field. This reframing is
what makes the physics prior usable here at all, and it generalizes: many
``physics-informed'' problems on discrete or combinatorial domains are better
served by physics-derived \emph{supervision} than by PINN-style residual losses.

\section{Related Work}
\textbf{Synthesizability scoring.} SAScore combines a fragment-contribution score
(rare fragments penalized) with a complexity penalty (rings, stereocenters,
macrocycles); it is fast, closed-form, and has been the de facto baseline for
fifteen years. Because it is a deterministic function of the molecular graph, it
is also exactly the kind of target a GNN can memorize---a fact we return to when
interpreting our in-distribution numbers. SCScore trains a network to score
complexity by ranking reaction-corpus products above their reactants. RAscore
classifies the binary outcome of the AiZynthFinder retrosynthesis planner
\citep{genheden2020}: a molecule is ``synthesizable'' if a route to purchasable
starting materials is found within a search budget. DeepSA trains a GNN on
SAScore-derived labels and matches or exceeds SAScore on held-out sets; it is the
closest published baseline to our Phase 1. More recent learned scorers move to
graph models: GASA \citep{gasa2022} casts easy/hard synthesizability as a
graph-attention classification task and explicitly targets generalization beyond
the training corpus, while SYBA \citep{syba2020} is a fragment-based Bayesian
easy/hard classifier; both are natural points of comparison for our GINE baseline.
The deployment motivation is sharpened by Gao and Coley \citep{gao2020}, who show
that generative models routinely propose molecules that standard scorers and
retrosynthesis tools flag as non-synthesizable---precisely the setting in which an
OOD-robust filter matters. None of these methods use physical priors---they are all
data-driven, and inherit the biases of their training corpora, which is exactly the
OOD failure mode we target.

\textbf{Auxiliary losses and physics in molecular ML.} Multi-task learning is
well established in molecular property prediction \citep{yang2019,stokes2020}, with
auxiliary signals that are usually other \emph{biological} property labels (LogP,
molecular weight, QED) or self-supervised graph tasks. Closest to us, MolBERT
\citep{fabian2020} pretrains a molecular language model with an auxiliary objective
that predicts a large bank of RDKit-computed physicochemical descriptors---the
Bertz index among them---and Dey and Ning \citep{dey2024} jointly train a GNN with
self-supervised auxiliary tasks, reconciling conflicting gradients with a
rotation-based surgery scheme. We do \emph{not} claim that supervising a molecular
net with a cheap, graph-computable descriptor is itself new---it is not. Our
narrower contribution is to (i) reduce the auxiliary signal to one or two
\emph{physically motivated} targets (topological complexity, force-field strain)
used as a lightweight regularizer rather than a broad descriptor-prediction
pretraining objective, and (ii) evaluate that regularizer specifically for
\emph{out-of-distribution} synthesizability filtering. The ``physics as
supervision, not PINN residual'' stance is likewise established: physics-guided
loss terms are the core of theory-guided data science \citep{karpatne2017}, and
Takamoto et al.\ \citep{takamoto2025} recently added physics-derived
weak-supervision losses to interatomic potentials to improve robustness under
shift; we transplant that paradigm to a discrete-graph classification filter. Our
proposal also differs from the input-based physics line, which feeds force-field-
or 3D-coordinate-derived information to the model as \emph{inputs} (SchNet
\citep{schutt2017}, DimeNet \citep{gasteiger2020}, MACE \citep{batatia2022}): those
models \emph{consume} physics through 3D coordinates, whereas we \emph{supervise} a
2D-graph model with a target that summarizes physical strain, keeping the
architecture and inputs identical to a standard, cheap GNN pipeline. This makes the
prior a drop-in addition rather than an architectural commitment, and keeps
inference cost identical to the baseline.

\textbf{Out-of-distribution generalization of GNNs.} A growing literature
documents that GNNs latch onto spurious, corpus-specific correlations and degrade
under distribution shift, and proposes remedies that decorrelate graph
representations \citep{li2022ood} or isolate stable causal subgraphs
\citep{fan2024}. Benchmarks such as GOOD \citep{good2022} and DrugOOD
\citep{drugood2023} formalize covariate- and concept-shift splits (scaffold, size,
assay) for exactly this purpose; our train-on-drug-like / test-on-natural-product
protocol is a source-based covariate-shift instance in that family, not a new split
type. We share the OOD diagnosis but take a deliberately lighter remedy: instead of
a new OOD-specific architecture or a regularizer \emph{learned} from the data, we
ask whether cheap, \emph{fixed} physical priors supplied as auxiliary supervision
buy OOD robustness on an off-the-shelf GNN.

\section{Method}
\textbf{Backbone.} A 4-layer Graph Isomorphism Network with edge features (GINE,
\citealp{hu2020}): 40-dim atom features (type, degree, formal charge,
hybridization, total hydrogens, aromaticity, ring membership), 6-dim bond features
(type, conjugation, ring membership), hidden width 128, dropout 0.1, sum pooling.
Two heads share the graph-level embedding: a classification head $f_{\text{cls}}$
producing the easy/hard logit $z$, and an auxiliary regression head
$f_{\text{aux}}$ producing a scalar. The baseline supervises only $f_{\text{cls}}$
with binary cross-entropy; $f_{\text{aux}}$ is an architectural placeholder until
an auxiliary loss is switched on. Crucially, the auxiliary heads are used only at
\emph{training} time; inference uses $f_{\text{cls}}$ alone, so the deployed filter
is identical in cost to the baseline.

\textbf{Topological complexity loss.} We supervise $f_{\text{aux}}$ with the Bertz
molecular complexity index \citep{bertz1981} (computed in closed form by RDKit), a
Huber regression on $z$-scored targets:
$\mathcal{L}_{\text{complexity}}=\mathrm{Huber}\!\big(f_{\text{aux}}(g),\,
\mathrm{BertzCT}(\text{mol})\big)$. The intent is to force the shared encoder to
represent calibrated structural complexity rather than collapse to features that
merely separate the training distribution.

\textbf{Strain-energy soft penalty.} We compute MMFF94 \citep{halgren1996} energy
per heavy atom $E_{\text{strain}}\!\ge\!0$ from a single ETKDG conformer, and
penalize confident ``easy'' predictions on strained molecules:
$\mathcal{L}_{\text{strain}}=\mathbb{E}\big[\sigma(z)\cdot
E_{\text{strain}}(\text{mol})\big]$, where $\sigma$ is the sigmoid. The penalty is
asymmetric---it does not penalize ``hard'' predictions---and acts as a
Lagrangian-style soft constraint: high strain usually means hard to synthesize
(cubane is the famous exception), but the model may override the prior when the
data are unambiguous.

\textbf{Combined objective.}
$\mathcal{L}=\mathrm{BCE}(z,y)+\lambda_c\mathcal{L}_{\text{complexity}}
+\lambda_s\mathcal{L}_{\text{strain}}$, with the baseline at
$\lambda_c=\lambda_s=0$. We use $(\lambda_c,\lambda_s)=(0.5,0.1)$, chosen so the
aux term is $\sim$10--30\% of the BCE magnitude. The two losses are always ablated
separately and jointly, never bundled into a single uninterpretable ``physics
loss.''

\section{Experimental Setup}\label{sec:setup}
\begin{figure}[t]
\centering
\includegraphics[width=0.78\linewidth]{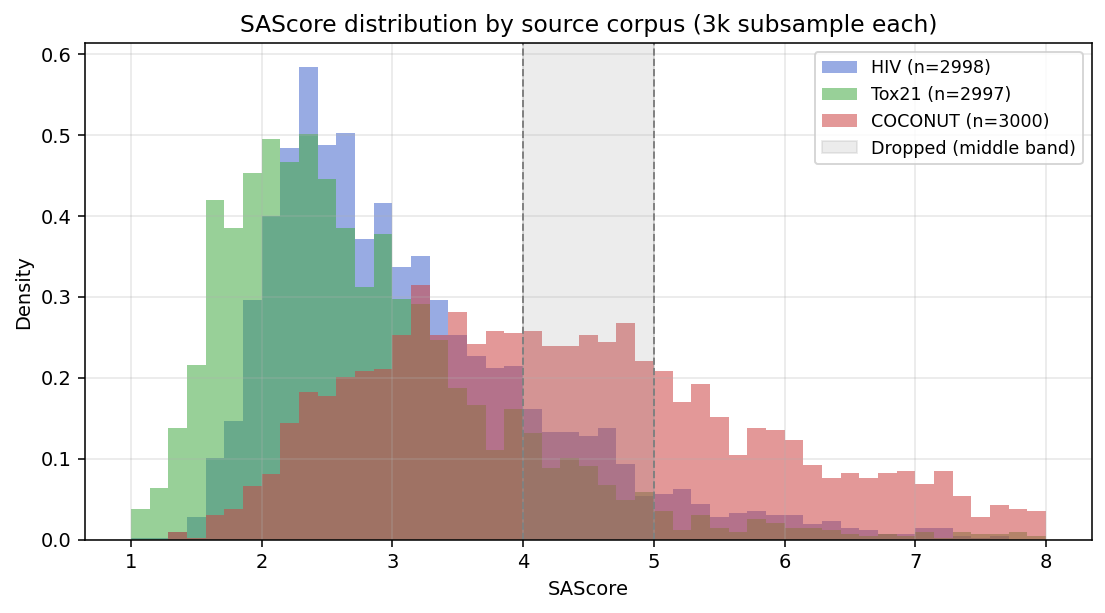}
\caption{SAScore distribution per source (subsampled). COCONUT natural products
(right-skewed) are systematically ``harder'' by SAScore than HIV/Tox21 drug-like
molecules; the shaded band ($4\le\mathrm{SAScore}\le5$) is dropped at labeling.
This distributional gap motivates the single-source OOD protocol.}
\label{fig:sascore}
\end{figure}

\textbf{Data.} No single public corpus holds both drug-like and complex
natural-product molecules at scale, so we combine three public sources: HIV
(41{,}127 SMILES) and Tox21 (7{,}831) from MoleculeNet via DeepChem, and a
30{,}000-molecule uniform subsample of COCONUT natural products
\citep{sorokina2021}. Labels follow the DeepSA bootstrap convention via SAScore:
$<\!4\Rightarrow$ easy (label 1), $>\!5\Rightarrow$ hard (label 0), ambiguous
middle band dropped. We relax the hard threshold from the conventional 6 to 5
because the strict split left only $\sim$1{,}158 hard molecules; we note this
departure and revisit it in the limitations. After dedup and labeling:
\textbf{65{,}177 molecules} (53{,}159 easy / 12{,}018 hard, an 82/18 split).
Figure~\ref{fig:sascore} shows the per-source SAScore distributions: COCONUT skews
harder than the drug-like corpora, with substantial overlap.

\textbf{Ablation subset.} The aux-loss ablation uses a stratified
10{,}000-molecule subset (5k easy / 5k hard) so the per-molecule MMFF94 precompute
is tractable ($\sim$14 min on 10 CPU workers). Bertz is computed for all
molecules; MMFF94 strain is available for 73.2\% (molecules $>$50 heavy atoms are
skipped because ETKDG can hang on large peptides/macrocycles; their strain target
is set to 0, contributing nothing to the penalty). This skip is markedly more
common on the COCONUT OOD \emph{test} corpus---only 59.6\% of test molecules carry
a non-zero strain target (40.4\% are zeroed as oversized natural products)---so the
+strain variant's OOD gain operates largely through the training-time
representation constraint rather than a strain signal available at test time.

\textbf{OOD protocol.} The core experiment trains on HIV+Tox21 \emph{only}
(drug-like) and tests on COCONUT \emph{only} (natural products), isolating whether
the model learned a transferable representation or merely a corpus-classifier.
Each variant trains 30 epochs (AdamW, lr $10^{-3}$, weight decay $10^{-5}$, batch
128); the best-validation-AUC epoch selects the checkpoint. We repeat over 5 seeds
(each seeding the train/val split, weight initialization, and dropout---all RNGs
are seeded for reproducibility) and report the mean delta against the same-seed
baseline with a 95\% paired bootstrap CI. \emph{The bootstrap unit is the seed.}
For each aux variant we form the five per-seed paired deltas (its OOD AUC minus the
\emph{same-seed} baseline's), resample those five deltas with replacement
$10{,}000$ times, and take the $2.5$/$97.5$ percentiles of the resampled mean
delta. We resample seeds, not test molecules within a seed, so the interval
reflects run-to-run variability (initialization and the train/val split) rather
than within-corpus sampling noise. We stress that with only five seeds this
bootstrap is \emph{coarse}: the resampled mean is supported on a small set of
distinct values, so the interval should be read as indicative rather than exact. We
therefore corroborate every CI with the per-seed sign counts
(Sec.~\ref{sec:results}), which make no resampling assumption, and we flag a larger
seed grid as the obvious way to tighten the estimate.

\section{Pre-registered Claims}\label{sec:claims}
To guard against post-hoc storytelling, we fixed the claims and the success
criteria \emph{before} running the multi-seed ablation:\footnote{The success
criterion (paired bootstrap $p<0.05$ without in-distribution degradation) was
recorded in our project roadmap, timestamped in the anonymized code release,
before the confirmatory multi-seed run. Because an exploratory single-seed run
preceded it (see the cautionary tale in the Discussion), we scope this as
pre-registration of the \emph{confirmatory} analysis, not of the initial
hypothesis.}
\begin{itemize}
\item \textbf{Claim 1 (necessary, not sufficient).} The GINE baseline reproduces a
DeepSA-class in-distribution classifier (ROC-AUC in the DeepSA-reported range,
$\ge 0.85$). Without this, no ablation built on the baseline is meaningful.
\item \textbf{Claim 2 (main).} At least one of $\{$+complexity, +strain, +both$\}$
improves OOD ROC-AUC over the baseline by a statistically significant margin
(paired bootstrap, $p<0.05$, i.e.\ 95\% CI excluding zero) \emph{without}
degrading in-distribution AUC by more than 0.01.
\end{itemize}
We committed to publishing the result either way: had only Claim 1 held, we would
have reported a null result for physics-aware regularization, which the field
needs (the default assumption that physical priors always help is itself worth
falsifying). As Sec.~\ref{sec:results} shows, both claims are met---and, in fact,
\emph{all three} aux variants satisfy Claim 2, a stronger outcome than the
disjunction we pre-registered.

\section{Results}\label{sec:results}

\subsection{In-distribution baseline (plumbing, not a finding)}
\begin{figure}[t]
\centering
\includegraphics[width=\linewidth]{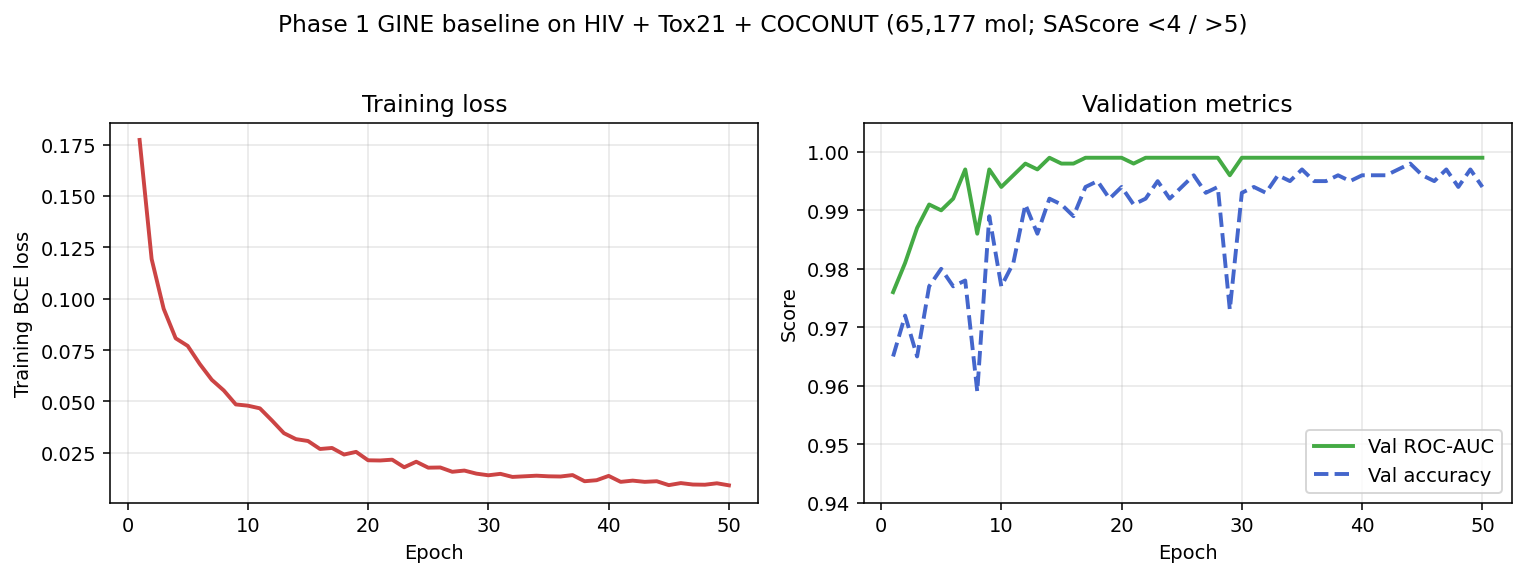}
\caption{Phase 1 baseline training loss (left) and validation ROC-AUC (right).
Validation AUC exceeds 0.99 by epoch 4 and converges by $\sim$epoch 15; validation
accuracy lags slightly because of the 82/18 class imbalance.}
\label{fig:phase1}
\end{figure}

On the full 65{,}177-molecule corpus the GINE baseline reaches the numbers in
Table~\ref{tab:phase1} and the curves in Figure~\ref{fig:phase1}. We frame the
near-perfect AUC as \emph{baseline reproduction}, not a research result, for three
reasons. \textbf{(1)~SAScore is closed-form in the molecular graph}---a weighted
sum of fragment contributions plus a complexity penalty---so a sufficiently
expressive GNN will reproduce it in the limit, matching the $\ge\!0.95$ AUC DeepSA
reports under the same convention. \textbf{(2)~The dropped middle band erases the
boundary}: no labeled molecule has a true SAScore within 1.0 of the decision
threshold, so the classifier never sees the hard boundary cases that would expose
its calibration. \textbf{(3)~A source-corpus confound}: HIV/Tox21 have mean
SAScore $\approx 2.9$--$3.1$ and COCONUT $\approx 4.4$ (Fig.~\ref{fig:sascore}), so
part of the signal is corpus discrimination rather than synthesizability.
\textbf{(4)~The splits share no molecule}: train, validation, and test are a random
partition of a corpus carrying no duplicate SMILES (only four of the 65{,}177
molecules share a canonical SMILES form), so no molecule appears in more than one
fold and the $1.000$ reflects reproduction of a closed-form labeling function, not
train/test leakage. The interesting question is therefore not in-distribution
accuracy but OOD behavior; Claim 1 is satisfied and we move on.

\begin{table}[t]
\centering\small
\begin{tabular}{lc}
\toprule
Metric & Value \\
\midrule
Best epoch (by val AUC) & 14 \\
Best validation ROC-AUC & 0.999 \\
\textbf{Test ROC-AUC} & \textbf{1.000} \\
Test accuracy & 0.997 \\
Training wall time (50 epochs, 1 GPU) & $\approx$4 min \\
\bottomrule
\end{tabular}
\caption{Phase 1 baseline on the full 65k corpus. Read as a reproduction of a
closed-form labeling function (see text), not as evidence of learned
synthesizability beyond SAScore.}
\label{tab:phase1}
\end{table}

\subsection{In-distribution ablation}
\begin{figure}[t]
\centering
\includegraphics[width=\linewidth]{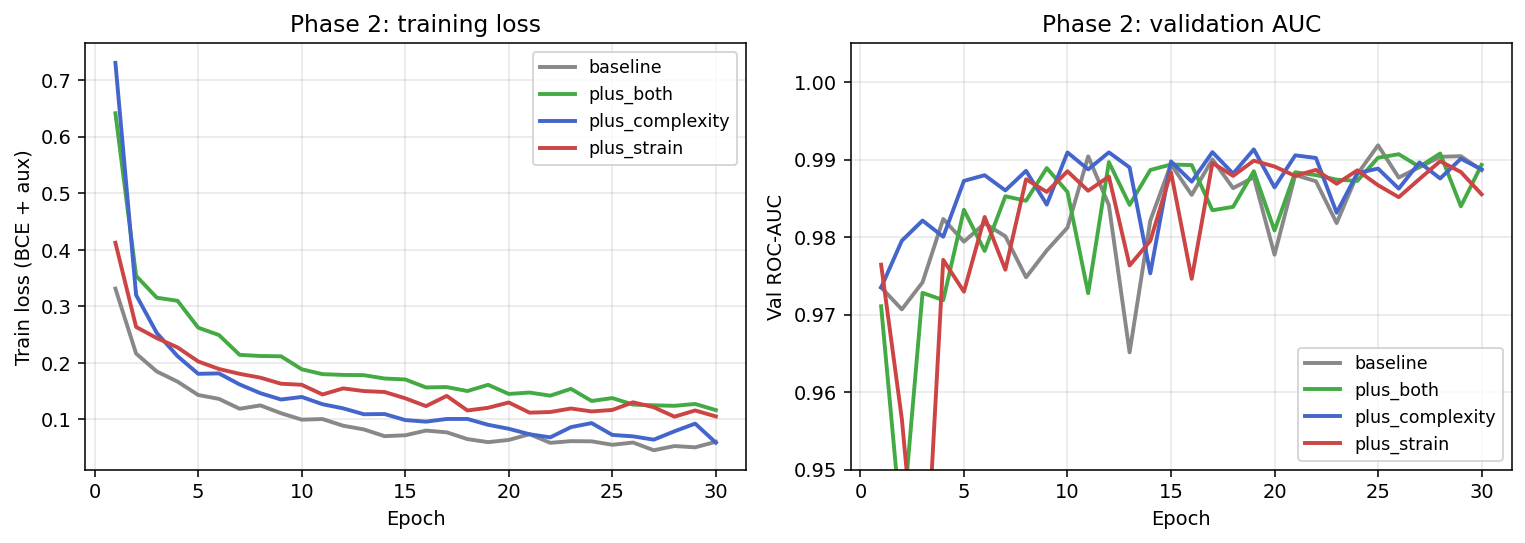}
\caption{In-distribution training loss (left; the strain/combined variants carry a
constant aux offset) and validation ROC-AUC (right; all variants converge to
$\sim$0.99). The variants are indistinguishable in-distribution.}
\label{fig:phase2}
\end{figure}

On a random 80/10/10 split of the 10k subset, all four variants reach essentially
the same test AUC (Table~\ref{tab:phase2}, Fig.~\ref{fig:phase2}): they lie within
$\sim$0.002 AUC of one another. Aux losses neither help nor hurt
in-distribution---consistent with the closed-form-label framing---which both
confirms they are not destructive as a drop-in regularizer (the second half of
Claim 2) and shows that any benefit must be sought OOD.

\begin{table}[t]
\centering\small
\begin{tabular}{lccccc}
\toprule
Variant & $\lambda_c$ & $\lambda_s$ & Best Val AUC & Test AUC & Test Acc \\
\midrule
Baseline GINE        & 0.0 & 0.00 & 0.9919 & 0.9963 & 0.9700 \\
+ Bertz complexity   & 0.5 & 0.00 & 0.9913 & 0.9948 & 0.9480 \\
+ MMFF94 strain      & 0.0 & 0.10 & 0.9899 & 0.9956 & 0.9740 \\
+ Both               & 0.5 & 0.10 & 0.9908 & 0.9965 & 0.9690 \\
\bottomrule
\end{tabular}
\caption{In-distribution ablation (random 80/10/10 split of the 10k subset). All
four variants are within run-to-run noise; the aux losses do not move the iid
number.}
\label{tab:phase2}
\end{table}

\subsection{Out-of-distribution ablation (main result)}
\begin{figure}[t]
\centering
\includegraphics[width=\linewidth]{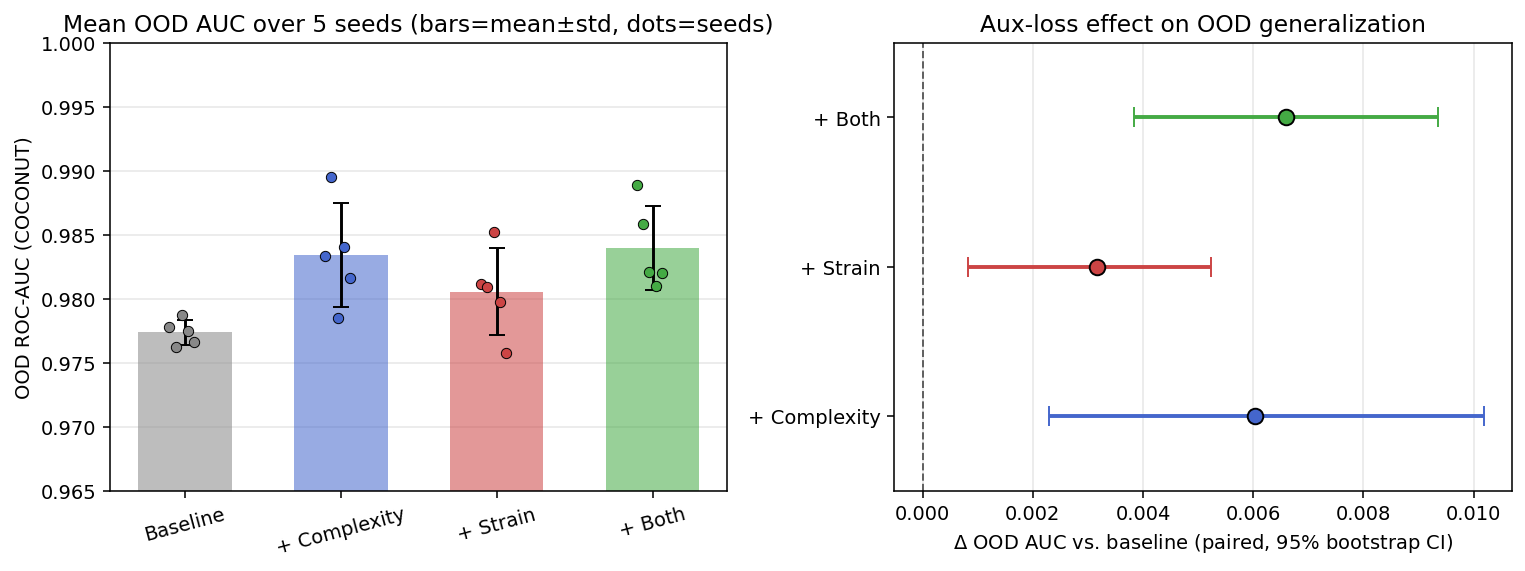}
\caption{\textbf{Left:} mean OOD ROC-AUC per variant (bars = mean $\pm$ std over 5
seeds; dots = individual seeds); all three aux variants sit above the baseline.
\textbf{Right:} paired $\Delta$ vs.\ baseline with 95\% bootstrap CIs; all three
intervals lie entirely right of zero.}
\label{fig:ood}
\end{figure}

Table~\ref{tab:ood} and Figure~\ref{fig:ood} report the 5-seed OOD result.
\textbf{All three physics-aware variants significantly improve OOD AUC} over the
baseline; every paired bootstrap CI excludes zero, and the combination is best.
Claim 2 is satisfied---by all three variants, not just one. The per-seed signs
corroborate the intervals (Table~\ref{tab:perseed}): +both improves OOD AUC over
the same-seed baseline on all five seeds, +complexity on four of five, and +strain
on four of five---a sign count we report alongside the CIs because it is read
directly off the per-seed table and needs no resampling assumption.

\begin{table}[t]
\centering\small
\begin{tabular}{lccccc}
\toprule
Variant & $\lambda_c$ & $\lambda_s$ & Mean OOD AUC & $\Delta$ vs base & 95\% CI \\
\midrule
Baseline GINE          & 0.0 & 0.00 & 0.9774 (\,$\pm$0.0010) & ---      & --- \\
+ Bertz complexity     & 0.5 & 0.00 & 0.9834 (\,$\pm$0.0040) & $+0.0060$ & $[+0.0023,+0.0102]$ \\
+ MMFF94 strain        & 0.0 & 0.10 & 0.9805 (\,$\pm$0.0034) & $+0.0032$ & $[+0.0008,+0.0052]$ \\
\textbf{+ Both}        & 0.5 & 0.10 & \textbf{0.9840} (\,$\pm$0.0033) & $\mathbf{+0.0066}$ & $\mathbf{[+0.0038,+0.0093]}$ \\
\bottomrule
\end{tabular}
\caption{Multi-seed (5 seeds) OOD ablation: train on drug-like HIV+Tox21, test on
COCONUT natural products. Every aux variant's paired bootstrap CI excludes zero.
Per-seed values in Appendix~\ref{app:perseed}.}
\label{tab:ood}
\end{table}

\textbf{External reference points.} To situate these numbers we score the
\emph{same} COCONUT OOD test set with several external scorers
(Table~\ref{tab:external}). SAScore itself reaches AUC $1.000$, but this is
\emph{definitional}: our labels are thresholded SAScore, so SAScore is the labeling
function and its perfect score is circular rather than an independent
comparison---which is exactly why a deployable filter cannot simply call SAScore
and must instead learn a representation that transfers. More informative is
SCScore \citep{coley2018}, an \emph{independent} learned scorer trained to rank
reaction-corpus products by synthetic complexity rather than on SAScore labels: on
the same set it reaches only AUC $0.656$. We are careful not to read this as
SCScore being ``worse''---it targets a related-but-distinct notion of
synthesizability---but as direct evidence that two learned synthesizability scorers
disagree substantially on natural products, i.e.\ there is no shared ground truth,
which is precisely the corpus-dependence our OOD protocol probes and the reason we
frame our result as improved \emph{agreement with SAScore} rather than with truth.
Two naive structural proxies, molecular weight and heavy-atom count, reach AUC
$0.785$ and $0.782$, showing that raw molecular size already carries substantial
signal on this drug-like$\rightarrow$natural-product split. Our GINE baseline
($0.977$) and physics-aware variants ($0.984$) sit above every independent
reference point---unsurprising, since they are trained on the SAScore labels, so
this contextualizes rather than validates the absolute numbers. A second
independent GNN scorer (GASA \citep{gasa2022}) would further tighten the picture; we
leave that to future work.

\begin{table}[t]
\centering\small
\begin{tabular}{lc}
\toprule
Scorer on COCONUT OOD test (5{,}026 mol) & ROC-AUC \\
\midrule
SAScore (= labeling function; circular) & 1.000 \\
SCScore (independent learned scorer) & 0.656 \\
Molecular weight (naive size proxy) & 0.785 \\
Heavy-atom count (naive size proxy) & 0.782 \\
GINE baseline (ours, trained on SAScore labels) & 0.977 \\
\textbf{+ Both physics-aware losses (ours)} & \textbf{0.984} \\
\bottomrule
\end{tabular}
\caption{External reference points on the COCONUT OOD test set (1{,}326 easy /
3{,}700 hard), scored against the SAScore-derived labels. SAScore is the labeling
function, so its AUC of $1.000$ is true by construction (not an independent
baseline). SCScore---an independent learned scorer not trained on SAScore---reaches
only $0.656$, evidence that learned synthesizability scorers disagree on natural
products. Naive size proxies reach $\sim$0.78. Our filters are trained on the
SAScore labels, so their high agreement contextualizes rather than validates the
absolute numbers.}
\label{tab:external}
\end{table}

\section{Discussion}
\textbf{Why the effect appears only OOD.} The four variants are statistically
indistinguishable in-distribution (Table~\ref{tab:phase2}) yet separate cleanly
under corpus transfer (Table~\ref{tab:ood}). The natural reading is that on the
training distribution the BCE signal already lets a high-capacity GNN recover
SAScore, so the auxiliary targets are redundant; under distribution shift,
however, the auxiliary supervision constrains the shared representation toward
features (calibrated complexity, strain sensitivity) that remain informative on a
corpus the model never trained on. This is the paper's central methodological
point: \emph{a filter meant to sit downstream of a generative model must be
evaluated OOD}, because in-distribution metrics hide precisely the differences
that matter in deployment. The point is not specific to synthesizability: any
learned filter applied to generative output inherits this evaluation requirement.

\textbf{What we do and do not claim.} We claim that cheap, closed-form physical
targets used as auxiliary supervision consistently and significantly improve OOD
generalization of a GNN synthesizability filter on a drug-like$\rightarrow$
natural-product split. Because the labels are an SAScore proxy, this is, strictly,
improved \emph{agreement with SAScore} on a shifted corpus---a
necessary-but-not-sufficient proxy for true synthesizability, as the Limitations
detail. We do \emph{not} claim a large effect (the gains are
sub-1\% absolute on an already-high 0.977 baseline, though +0.006 is a $\sim$26\%
reduction of the $(1{-}\text{AUC})$ error gap), a universal one (we test one OOD
axis and one weight configuration), or a reliable ranking among the three variants
(with five seeds the +complexity and +both intervals overlap substantially). The
robust, weaker statement is the one we stand behind. Because ROC-AUC is
near-saturated at a 0.977 baseline, we frame the effect against the
$(1{-}\text{AUC})$ error gap and as per-seed paired deltas rather than raw AUC; a
threshold-free precision--recall analysis would sharpen the comparison but needs
per-sample scores we did not retain across the seed grid.

\textbf{Deployment guidance.} Because the labels are an SAScore proxy and the
effect is modest, the filter should be used as \emph{one signal among several},
not a hard gate---an over-trusted filter could systematically down-rank
legitimately novel-but-makeable scaffolds. The value of the physics-aware variant
is specifically that it degrades less under the distribution shift a generative
model induces.

\textbf{A cautionary tale about single-seed findings.} An earlier single-seed run
of this exact ablation produced a confident, qualitatively different story: strain
alone appeared to \emph{hurt} OOD AUC ($-0.0125$), complexity looked neutral, and
only the combination helped---a ``non-monotone interaction'' narrative. That
pattern did not survive multi-seed evaluation. It was an artifact of uncontrolled
weight initialization: the original code seeded only the data split, not the model
parameters. After seeding all RNGs and repeating over five seeds, every variant is
positive and the non-monotone structure vanishes (per-seed numbers in
Appendix~\ref{app:perseed} show the effect is consistent-on-average but not
unanimous---complexity dips to $-0.0002$ on one seed, strain to $-0.0009$ on
another, which is exactly why a paired bootstrap CI, not a single run, is the
right instrument). We report this prominently because it is a concrete, in-domain
example of how an AI-assisted research loop can manufacture a compelling but false
narrative from a single sub-1\%-AUC run---and a small argument for multi-seed
confidence intervals as a default discipline in AI-assisted science.

\section{Limitations and Threats to Validity}
\textbf{Construct validity. SAScore labels are a proxy}, not ground truth; a model
trained on them is at best an SAScore-emulator on the training distribution, which
is the reason we test OOD rather than trusting in-distribution accuracy.
\textbf{Bertz is hand-engineered}; a model that perfectly predicts Bertz has not
necessarily learned anything useful---the aux loss matters only through its
downstream OOD effect, which is what we measure, not aux-head accuracy.

\textbf{Internal validity. Threshold relaxation} ($>\!5$ rather than the
conventional $>\!6$) narrows the gap between class supports and makes the iid task
slightly easier; a follow-up should reproduce both thresholds.
\textbf{Single-conformer MMFF94} approximates true ensemble strain; the exact
quantity depends on the conformer distribution, and MMFF94 fails outright on
organometallics and some non-organic elements (those rows are dropped, which could
itself bias the strain signal).

\textbf{External validity. The effect is small and tested on one OOD axis} with
one weight configuration and five seeds. The five-seed paired bootstrap is
correspondingly coarse and should be read as indicative; we lean on the per-seed
sign counts alongside it. Our external baselines are one independent
learned scorer (SCScore) plus SAScore (circular here) and size proxies; a second
independent learned GNN scorer (GASA) on the same split would further strengthen
the comparison. Additional axes (macrocycles, peptidomimetics,
organometallics, and especially \emph{generator outputs}, the deployment target), a
$(\lambda_c,\lambda_s)$ grid sweep, and more seeds would sharpen---or qualify---the
claim. We pre-registered and report the result at this scope rather than
over-generalizing.

\section{Conclusion}
Synthesizability prediction is a setting where physics-grounded bias should enter
as auxiliary regularization on a strong GNN baseline rather than as a PINN
(mathematically inappropriate for discrete graphs). Across 5 seeds, all three
physics-aware aux variants significantly improve OOD AUC over a purely data-driven
baseline, with the combination best ($+0.0066$, 95\% CI $[+0.0038,+0.0093]$) and
the effect visible only OOD. The improvements are modest but robust. Pure
statistical filters inherit their training corpus's biases; cheap physics-aware
regularization is a drop-in step toward the OOD-robust filters that generative
drug-design pipelines actually need. Just as importantly, our retracted
single-seed finding is a reminder that small-effect claims demand multi-seed
confidence intervals before they are trusted---a discipline we hope AI-assisted
research adopts by default.

\appendix
\section{Per-seed OOD results}\label{app:perseed}
Table~\ref{tab:perseed} gives the per-seed OOD ROC-AUC behind
Table~\ref{tab:ood}. The effect is consistent on average but not unanimous per
seed: +complexity is marginally negative on seed 2 ($-0.0002$) and +strain on
seed 4 ($-0.0009$); +both is positive on all five seeds. This non-unanimity is
why we report a paired bootstrap CI over seeds rather than a single run.

\begin{table}[h]
\centering\small
\begin{tabular}{lcccc}
\toprule
Seed & Baseline & + Complexity & + Strain & + Both \\
\midrule
0 & 0.9778 & 0.9833 & 0.9812 & 0.9889 \\
1 & 0.9762 & 0.9895 & 0.9809 & 0.9858 \\
2 & 0.9787 & 0.9785 & 0.9852 & 0.9821 \\
3 & 0.9775 & 0.9841 & 0.9797 & 0.9810 \\
4 & 0.9766 & 0.9816 & 0.9757 & 0.9820 \\
\midrule
Mean & 0.9774 & 0.9834 & 0.9805 & 0.9840 \\
\bottomrule
\end{tabular}
\caption{Per-seed OOD ROC-AUC (train HIV+Tox21, test COCONUT).}
\label{tab:perseed}
\end{table}

\end{document}